\documentclass[letterpaper]{article} 
\usepackage{aaai25}  
\usepackage{times}  
\usepackage{helvet}  
\usepackage{courier}  
\usepackage[hyphens]{url}  
\usepackage{graphicx} 
\usepackage{natbib}  
\usepackage{caption} 
\usepackage{algorithm}
\usepackage{algorithmic}

\usepackage{newfloat}
\usepackage{listings}
\DeclareCaptionStyle{ruled}{labelfont=normalfont,labelsep=colon,strut=off} 
\floatstyle{ruled}
\newfloat{listing}{tb}{lst}{}
\floatname{listing}{Listing}
\usepackage{makecell,multirow}
\usepackage{colortbl}
\usepackage{xcolor}
\usepackage{xfrac}
\usepackage{units}
\usepackage{enumitem}
\usepackage{amssymb}
\usepackage{array}
\usepackage{booktabs}

\usepackage{xspace}

\makeatletter
\DeclareRobustCommand\onedot{\futurelet\@let@token\@onedot}
\def\@onedot{\ifx\@let@token.\else.\null\fi\xspace}

\def\eg{\emph{e.g}\onedot} 
\def\ie{\emph{i.e}\onedot}

\makeatother

\title{Boosting Segment Anything Model Towards Open-Vocabulary Learning}
\author{
    Xumeng Han\textsuperscript{\rm 1}\thanks{This work was done during the internship in Huawei Inc.},
    Longhui Wei\textsuperscript{\rm 2}\footnotemark[2],
    Xuehui Yu\textsuperscript{\rm 1},
    Zhiyang Dou\textsuperscript{\rm 1},
    Xin He\textsuperscript{\rm 2},\\
    Kuiran Wang\textsuperscript{\rm 1},
    Yingfei Sun\textsuperscript{\rm 1},
    Zhenjun Han\textsuperscript{\rm 1}\thanks{Corresponding Authors.},
    Qi Tian\textsuperscript{\rm 2}
}
\affiliations{
    \textsuperscript{\rm 1}University of Chinese Academy of Sciences \\
    \textsuperscript{\rm 2}Huawei Inc.

    \{hanxumeng19, yuxuehui17, douzhiyang23, wangkuiran19\}@mails.ucas.ac.cn, \\ 
    \{weilh2568, whut.hexin\}@gmail.com,
    \{yfsun, hanzhj\}@ucas.ac.cn, tian.qi1@huawei.com
}

\usepackage{bibentry}

\begin{document}

\maketitle

\begin{abstract}
The recent Segment Anything Model (SAM) has emerged as a new paradigmatic vision foundation model, showcasing potent zero-shot generalization and flexible prompting. Despite SAM finding applications and adaptations in various domains, its primary limitation lies in the inability to grasp object semantics. 
In this paper, we present \textbf{Sambor} to seamlessly integrate SAM with the open-vocabulary object detector in an end-to-end framework. While retaining all the remarkable capabilities inherent to SAM, we boost it to detect arbitrary objects from human inputs like category names or reference expressions. 
Building upon the SAM image encoder, we introduce a novel SideFormer module designed to acquire SAM features adept at perceiving objects and inject comprehensive semantic information for recognition.
In addition, we devise an Open-set RPN that leverages SAM proposals to assist in finding potential objects.
Consequently, Sambor enables the open-vocabulary detector to equally focus on generalizing both localization and classification sub-tasks.
Our approach demonstrates superior zero-shot performance across benchmarks, including COCO and LVIS, proving highly competitive against previous state-of-the-art methods.
We aspire for this work to serve as a meaningful endeavor in endowing SAM to recognize diverse object categories and advancing open-vocabulary learning with the support of vision foundation models.
\end{abstract}

%

\begin{figure}
  \centering
  \includegraphics[width=1\columnwidth]{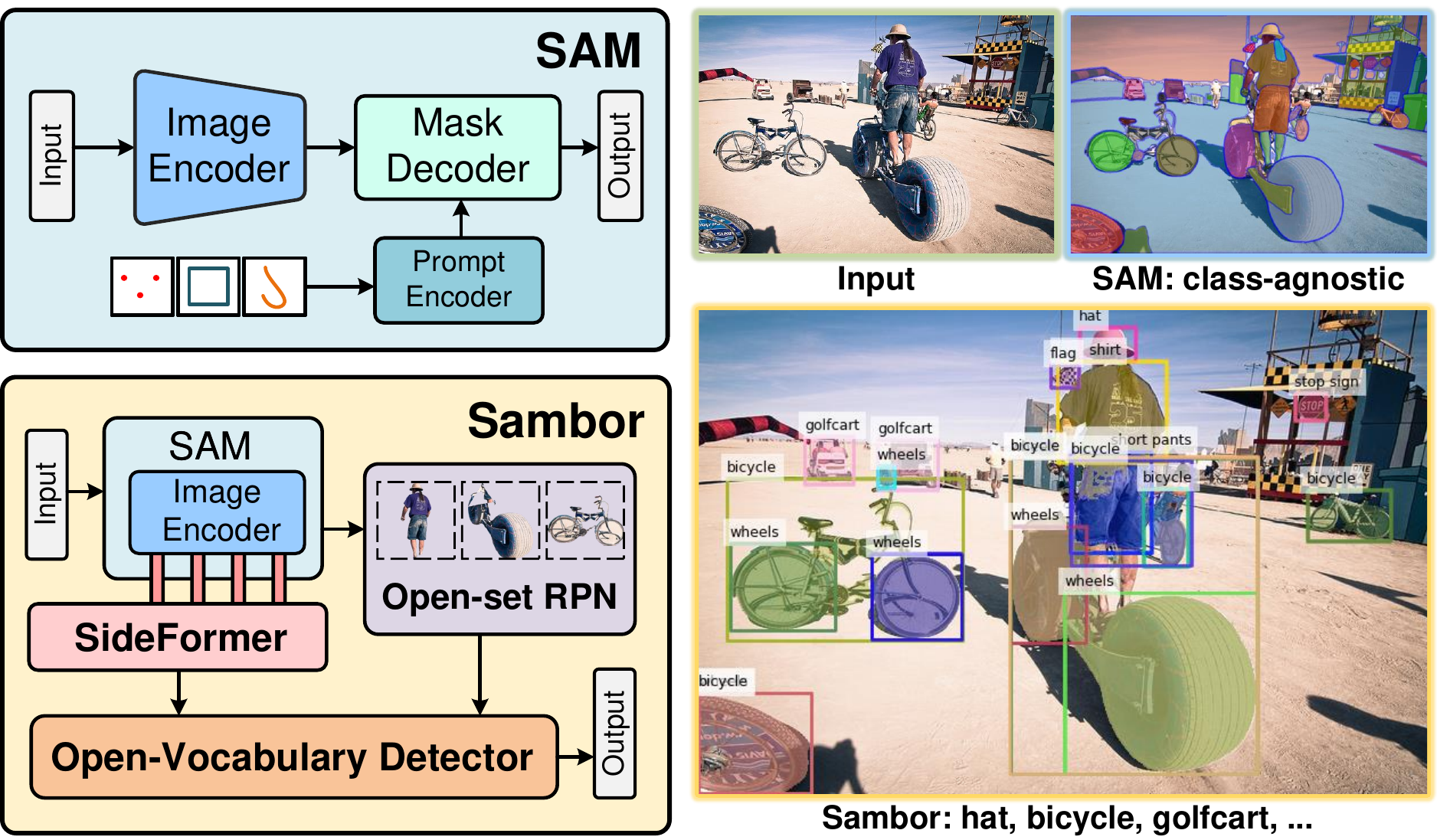}
  \caption{We develop an end-to-end open-vocabulary object detector called \textbf{Sambor}, building upon the vision foundation model SAM.
  Sambor enables SAM to recognize arbitrary object categories, bridging semantic gaps. It also leverages SAM's generalization and interactive capabilities to enhance zero-shot performance and extend versatility.
}
  \label{fig:intro}
\end{figure}

\section{Introduction}
\label{sec:intro}
Vision foundation models~\cite{CLIP,MAE,dinov2} serve as robust backbones that excel across a diverse spectrum of vision tasks. The recent Segment Anything Model (SAM)~\cite{SAM} has garnered widespread attention within the community as a foundational visual model for general image segmentation. Trained with billion-scale mask labels, it demonstrates impressive zero-shot segmentation performance, seamlessly applied across a variety of applications through simple prompting~\cite{grounded-sam,captionSAM}. While exhibiting outstanding performance, SAM is constrained to scenarios with class-agnostic applications, necessitating a more profound exploration to enhance its semantic understanding. In this work, we boost SAM towards open-vocabulary learning to detect objects of arbitrary categories, a paradigm commonly referred to as open-vocabulary object detection~\cite{OVR-CNN,survey}.

Recent works~\cite{ViLD,Regionclip,OWL-ViT,F-VLM,GLIP,DetCLIP,Grounding-DINO} commonly follow two lines to achieve open-vocabulary object detection. One seeks to expand the cognitive categories by transferring knowledge from pre-trained vision-language (VL) models~\cite{CLIP,ALIGN}. The other unifies the formulation of object detection and phrase grounding tasks, extending the available data scope from object detection to a diverse range of image-text pairs~\cite{CC3M,SBU,YFCC100M}. The utilization of large-scale data encourages the model to learn the feature alignment between object regions and language phrases.
Previous efforts have been built upon conventional object detection models, introducing the VL paradigm to extend the object detection classifier from the close-set to the open-set domain. With the emergence of the powerful vision foundation model SAM, its remarkable zero-shot localization capabilities and the flexibility for interactive prompting raise the prospect of elevating open-vocabulary object detection to new heights.

In this paper, we introduce \textbf{Sambor} (which stands for \textbf{SAM BO}oste\textbf{R}), as illustrated in Fig.~\ref{fig:intro}. Sambor is an end-to-end open-vocabulary object detector that seamlessly integrates all functionalities from the SAM, inheriting its powerful zero-shot generalization and flexible prompting.
We establish a ladder side transformer adapter, named SideFormer, on the frozen SAM image encoder.
It incorporates an extractor designed to integrate features from the image encoder, drawing object perception capabilities from SAM.
Subsequently, we devise an injector to augment the original features by introducing additional semantic information to assist category recognition.
In pursuit of the injected features imbued with rich semantics, CLIP~\cite{CLIP}, another vision foundation model, naturally emerges as our excellent choice, given it achieves outstanding zero-shot classification through VL alignment.
As a result, infusing CLIP visual features not only enhances semantic understanding but also narrows the gap with the text feature domain, providing convenience for the open-vocabulary object classifier~\cite{ViLD,F-VLM}.

Moreover, SAM exhibits the capacity to generate high-quality class-agnostic proposals. To fully leverage this advantage and further augment the zero-shot localization ability, we develop the Open-set RPN.
Specifically, Sambor adopts the two-stage object detection architecture~\cite{Faster}, decoupling the detector into a first stage dedicated to generating high-quality proposals to ensure sufficient recall and a second stage focused on open-vocabulary classification.
Open-set RPN complements the vanilla RPN~\cite{Faster} by introducing proposals oriented towards open-set scenarios. Thanks to the flexibility afforded by SAM, these additional proposals can be obtained through various prompts or the automatic mask generation pipeline.

We follow the GLIP~\cite{GLIP} protocol and conduct experiments to comprehensively evaluate the effectiveness of Sambor in open-vocabulary object detection. Benefiting from the effective designs, Sambor demonstrates superior open-vocabulary detection performance on COCO~\cite{COCO} and LVIS~\cite{LVIS} benchmarks.
It implies that we endow SAM with the capability to recognize arbitrary objects, boosting it to be more versatile.
From another perspective, incorporating SAM into an end-to-end framework provides greater versatility compared to current state-of-the-art open-vocabulary detectors~\cite{GLIP,Grounding-DINO,DetCLIP,DetCLIPv2}. For instance, it allows for seamless conversion of object detection results into instance segmentation or facilitates human-machine interaction through prompts. These capabilities are previously either unavailable or required the cascading of multiple models~\cite{grounded-sam}, introducing additional complexity and operational challenges.
Given these encouraging results, we aspire to endow the vision foundation model SAM with recognition capabilities to address a broader spectrum of applications and offer a potential way for the development of open-vocabulary object detection.

\section{Related Work}

\noindent
\textbf{General Object Detection,} a crucial computer vision task, consists of two sub-problems: finding the object (localization) and naming it (classification). 
Convolution-based detectors are typically divided into two-stage~\cite{Faster,maskrcnn,Cascade} or single-stage~\cite{yolo,focalloss,atss}, based on hand-crafted anchors or reference points.
Recent transformer-based methods~\cite{detr,deformable} try to formulate object detection as a set prediction problem.
These methods are constrained to predefined categories, whereas our approach aims to find and recognize objects of arbitrary category in an open domain.

\noindent
\textbf{Open-Vocabulary Object Detection}~\cite{OVR-CNN,survey} has emerged as a new trend for modern object detection, which aims to detect objects of unbounded concepts using a more universal and practical paradigm~\cite{detic,OWL-ViT,OV-DETR,mq-det,yolow}. Some of these methods~\cite{ViLD,Regionclip,DetPro,F-VLM,CoDet,CORA,BARON} leverage vision-language models (VLMs)~\cite{CLIP} to align information between regions and words through a multi-stage pre-training strategy. 
GLIP~\cite{GLIP} pioneers an alternative approach that transforms the detection data into a grounding format and introduces a fusion module for simultaneous learning of vision-language alignment and object localization. 
Compared to previous methods, we utilize SAM to facilitate the finding of potential objects in the open domain, mitigating the issue where VLMs primarily focus on handling the intricate alignment between regions and texts, thus lacking in localization generalization.

\noindent
\textbf{Segment Anything Model}~\cite{SAM} is an innovative image segmentation model trained on the dataset comprising over 1 billion masks, designed to robustly segment any object guided by diverse prompts.
Influenced by its development, the community is focusing on developing more versatile and functional segmentation models~\cite{X-Decoder,OpenSeeD,seem,semantic-sam}.
In contrast, our method focuses on detecting objects of arbitrary categories and learning rich semantics from detection and grounding data, making it more suitable for open-world scenarios. We utilize SAM's prompt functionality to obtain masks directly from predicted bounding boxes, thereby eliminating the need for training a mask head and overcoming limitations posed by segmentation data volume.


\begin{figure*}[t]
  \centering
  \includegraphics[width=0.95\textwidth]{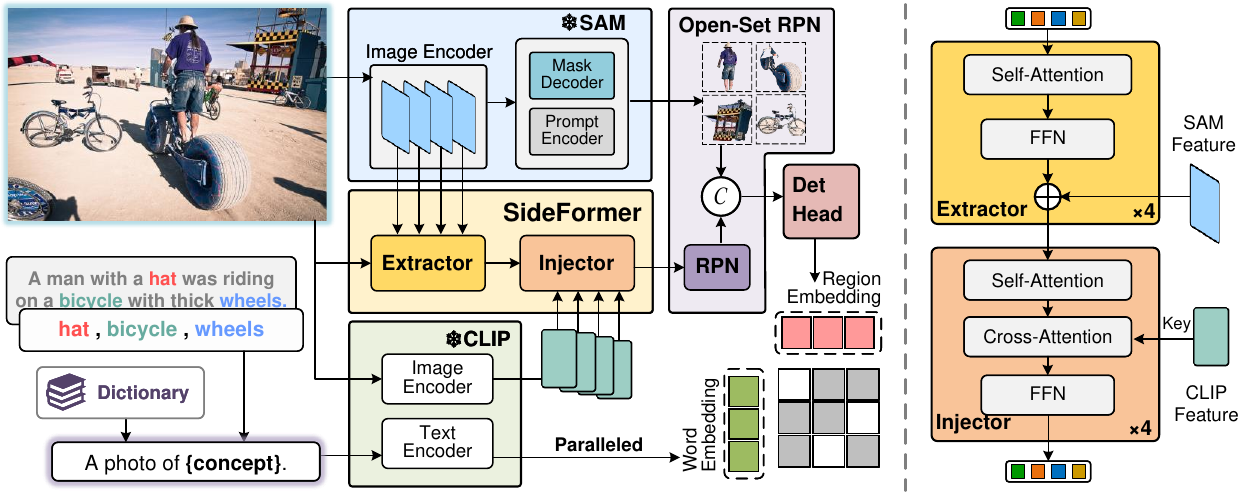}
  \caption{ \textbf{Overall architecture of Sambor. (Left)} 
We adopt the SAM image encoder as the backbone and construct a SideFormer module to extract features and inject CLIP visual information for enhancing semantic understanding.
Sambor is built upon a two-stage detector, with the first stage designed as an Open-set RPN that enhances the vanilla RPN using open-set proposals generated by SAM.
The second stage is equipped with a CLIP language branch for parallel concept encoding, thereby endowing the detector with open-vocabulary classification.
  \textbf{(Right)} The specific implementations of the extractor and injector. 
 }
  \label{fig:framework}
\end{figure*}

\section{Preliminaries}

\subsection{Open-Vocabulary Object Detection}
Given an image $\mathbf{I}\in\mathbb{R}^{3\times h\times w}$, object detection involves solving the two sub-problems of (1)~localization: find all objects with their location, represented as a box $\mathbf{b}_j$ and (2) classification: assign a class label $c_j\in\mathcal{C}^{\text{test}}$ to the $j$-th object. Here $\mathcal{C}^{\text{test}}$ is the class vocabulary provided by the user at test time. Traditional object detection considers $\mathcal{C}^{\text{test}}=\mathcal{C}^{\text{train}}$, where $\mathcal{C}^{\text{train}}$ denotes the vocabulary of detection dataset used during training.
Open-vocabulary object detection allows $\mathcal{C}^{\text{test}}\neq\mathcal{C}^{\text{train}}$. 
Taking GLIP~\cite{GLIP} as an example, it reformulates detection as a grounding task, aligning each visual region with corresponding class content in text prompts. Following CLIP~\cite{CLIP}, the model takes image-text pairs as input, extracting features from both modalities through dedicated encoders. By replacing the linear classification layer with a region-word matching dot product layer, it converts from a traditional detector to an open-vocabulary detector. 
However, previous open-vocabulary detection approaches primarily focus on improving classification across arbitrary categories, while their localization capability is still heavily reliant on bounding box annotations provided in detection datasets~\cite{DetCLIPv2}. 
In this paper, we compensate for the previously overlooked ability to find potential objects in the open domain, allowing the detector to equally focus on enhancing zero-shot generalization for object localization and classification.

\subsection{Vision Foundation Model}
\textbf{SAM}~\cite{SAM} is composed of three modules:
(1) Image encoder: a robust ViT-based~\cite{ViT} backbone excels at extracting features from high-resolution images. (2) Prompt encoder: encoding the interactive positional information from the input points, boxes, or masks.
(3) Mask decoder: a lightweight transformer-based~\cite{transformer} decoder efficiently translates the image and prompt embeddings into masks.

\noindent
\textbf{CLIP}~\cite{CLIP} leverages web-scale image-text pairs crawled from the Internet and simply aligning image features with text features via contrastive learning, delivering impressive results in zero-shot image classification.
In this paper, we choose the CNN-based CLIP model (\eg, RN50$\times$64~\cite{resnet}) over the ViT-based~\cite{ViT} one due to its superior compatibility with high-resolution image inputs~\cite{F-VLM}.

\section{Methodology}

An overview framework of Sambor is illustrated in Fig.~\ref{fig:framework}.
Our approach fully integrates SAM to leverage its exceptional zero-shot localization capabilities and the flexibility for interactive prompting, thus enhancing the performance and functionality of Sambor in open-world scenarios.
Specifically, we employ the SAM image encoder as the backbone and freeze the parameters during training. Due to the absence of a semantic prior in SAM features, the performance in category-aware tasks falls short of optimal. To address this issue, we introduce a ladder-side transformer adapter named SideFormer (Sec.~\ref{sec:sideformer}), conceptually similar to~\citet{LadderSide-Tuning,ViT-Adapter}. It is designed to extract features from the backbone and inject comprehensive semantic information from CLIP into them.

Subsequently, we construct an open-vocabulary object detector on the backbone, which is based on the two-stage ViTDet~\cite{ViTDet} with Cascade R-CNN~\cite{Cascade}.
For the two-stage detector, the first stage involves the RPN~\cite{Faster} for generating object proposals. We extend the RPN into an Open-set RPN (Sec.~\ref{sec:proposal}) by incorporating more zero-shot generalized proposals from SAM.
In the second stage, we equip it with a text encoder for open-vocabulary classification (Sec.~\ref{sec:classification}).
Therefore, we utilize these two stages to foster zero-shot generalization in both localization and classification for Sambor.



\subsection{SideFormer}
\label{sec:sideformer}
We design an extractor for acquiring SAM features adept at perceiving objects and an injector for absorbing knowledge from CLIP to enhance semantic understanding.

\begin{figure}[t]
  \centering
  \includegraphics[width=0.97\columnwidth]{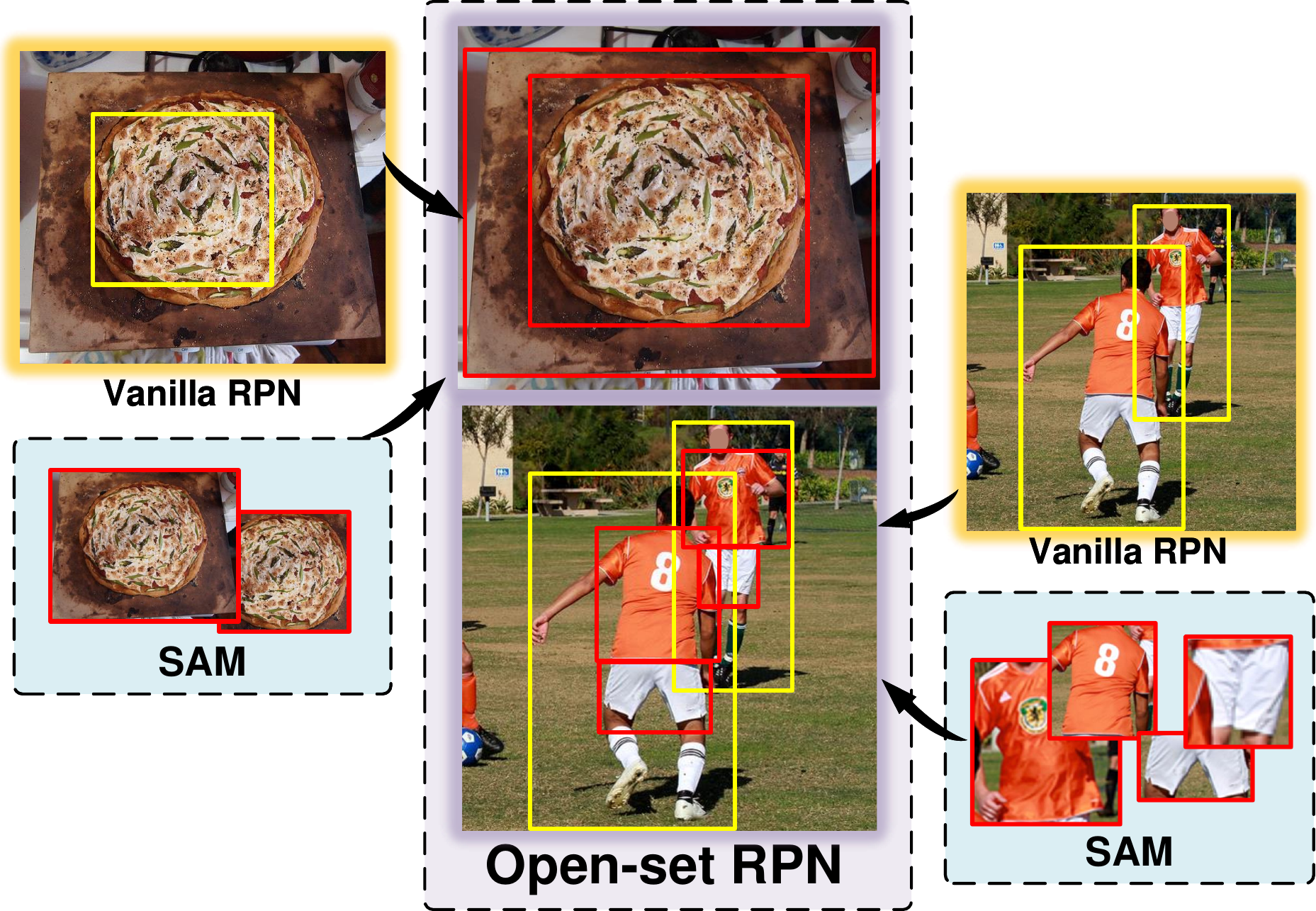}

  \caption{\textbf{An illustration of Open-set RPN.} We demonstrate two examples where SAM proposals effectively complement the vanilla RPN: \textbf{(Top-Left)} precise determination of object edge positions, and \textbf{(Bottom-Right)} clear capture of specific parts of an object, \eg, a person's clothing. 
  }
  \label{fig:openrpn}
\end{figure}

\noindent
\textbf{SAM Extractor.}
We adopt the patch embedding layer with a structure identical to that in SAM (without parameter sharing), initially projecting the input image into visual tokens.
The visual tokens are first summed with those from SAM and then incrementally fused with the deeper SAM features via a series of transformer layers~\cite{transformer}.
Following the ViTDet~\cite{ViTDet} architecture, SAM divides the ViT~\cite{ViT} into four blocks, applying global attention at the last layer in each block. 
In view of this, we design four corresponding transformer layers to extract features from each global attention output.

For each transformer layer, the input visual tokens $\mathcal{F}_{\mathrm{side}}\in\mathbb{R}^{\frac{hw}{16^2}\times D}$, where $D$ is the feature dimension, are encoded through a self-attention layer and a feed-forward network (FFN).
Subsequently, we use a direct summation to accomplish the fusion with the extracted SAM features $\mathcal{F}_{\mathrm{sam}}\in\mathbb{R}^{\frac{hw}{16^2}\times D}$. This process can be formulated as:
\begin{equation}
\hat{\mathcal{F}}_{\mathrm{side}}=\mathcal{F}_{\mathrm{side}}+\mathrm{Attn}(\mathrm{norm}(\mathcal{F}_{\mathrm{side}})),
\end{equation}
\begin{equation}
\mathcal{F}^{\prime}_{\mathrm{side}}=\mathcal{F}_{\mathrm{sam}}+\gamma\cdot(\hat{\mathcal{F}}_{\mathrm{side}}+\mathrm{FFN}(\mathrm{norm}(\hat{\mathcal{F}}_{\mathrm{side}}))),
\end{equation}where $\mathrm{norm}(\cdot)$ is LayerNorm~\cite{layernorm}. We apply a learnable vector $\gamma\in\mathbb{R}^D$ to modulate the transformer output, which is initialized with \textbf{0}. 
This strategy guarantees that the feature distribution of SideFormer commences from SAM without undergoing drastic alterations.

\noindent
\textbf{CLIP Injector.}
After integrating features extracted from SAM, we imbue them with semantically rich information from the CLIP~\cite{CLIP} visual encoder. The injector is also equipped with four transformer layers, supplemented by cross-attention modules for feature interaction.

We treat the input feature $\mathcal{F}^{\prime}_{\mathrm{side}}$ as the query, and the CLIP feature $\mathcal{F}_{\mathrm{clip}}$ as the key and value.
Initiate a self-attention encoding on the query first. Subsequently, we employ cross-attention on queries $\mathcal{F}^{\prime}_{\mathrm{side}}$ and $\mathcal{F}_{\mathrm{clip}}$, facilitating the assimilation of knowledge from CLIP. Finally, an FFN is appended, constituting the entirety of this transformer layer.
Eq.~\ref{eq:cross-attention} details the cross-attention module in the injector, while omitting the self-attention and FFN for brevity.
\begin{equation} \label{eq:cross-attention}
\mathcal{F}^{\,\ast}_{\mathrm{side}}=\mathcal{F}^{\prime}_{\mathrm{side}}+\gamma\cdot\mathrm{Attn}(\mathrm{norm}(\mathcal{F}^{\prime}_{\mathrm{side}}), \mathrm{norm}(\mathcal{F}_{\mathrm{clip}})),
\end{equation}where the learnable parameter $\gamma$ is introduced to likewise balance the injected CLIP features with the original inputs.

\subsection{Open-Set RPN}
\label{sec:proposal}
The primary aim of object detection is to thoroughly find potential objects, which implies the necessity for the RPN in the first stage to achieve a sufficiently high recall.
Hence, for the open-vocabulary object detector, possessing a robust RPN tailored for open-set domains is imperative.

To achieve this, we designed an Open-set RPN that integrates two proposal sources: the vanilla RPN~\cite{Faster} and SAM~\cite{SAM}.
We expect that the Open-set RPN will facilitate proposals from both sources to become complementary, aiming to find potential objects as comprehensively as possible.
The vanilla RPN learns from detection data and is adept at handling scenarios involving common objects. However, since the RPN trained on closed-set data needs to expand its generalization, we incorporate proposals from SAM as a valuable supplement.
We use the SAM head (consisting of the prompt encoder and the mask decoder) to generate masks and extract bounding boxes of these masks as object proposals.
Given SAM's robust zero-shot localization capability, its proposals can better cover areas that RPN has overlooked.
Fig.~\ref{fig:openrpn} primarily illustrates two possible scenarios to show how SAM proposals serve as supplements.
One is the enhanced precision in delineating the object contours, where SAM outperforms RPN in identifying the edge positions of overlapping objects or those with irregular shapes. The other scenario shows that SAM clearly captures parts of the whole object, \eg, a person's shirt and pants, areas where RPN tends to fall short on detail.
Nevertheless, more than relying solely on SAM is required; a trainable RPN is crucial as it reinforces handling common objects and compensates for deficiencies with small objects~\cite{SAM} of SAM. Please refer to Sec.~\ref{sec:ablation} and \ref{sec:analyses} for more detailed analysis and demonstrates.

Specifically, we leverage the automatic mask generation pipeline~\cite{SAM} with a point grid as prompts to predict mask proposals. 
Adjusting point density allows for control over the number of proposals, which in this paper defaults to a 32$\times$32 point grid as a balance between quantity and computational cost.
We adopt the straightforward NMS to merge two sets of proposals.
Given the domain gap in prediction scores between the two sets, it is unreasonable to apply NMS directly based on their scores.
Therefore, we first merge the proposals from RPN (which have already been de-duplicated) with those from SAM. Then, we perform NMS  (with the threshold set to 0.7) on SAM proposals using RPN boxes as references to filter out highly overlapping ones, thereby preserving areas not covered by RPN.

\setlength{\tabcolsep}{1.1pt}
\begin{table*}[tb]
\renewcommand\arraystretch{1.23}
\begin{center}
\resizebox{1\textwidth}{!}{
\begin{tabular}{lcccccccccc}
\specialrule{0.1em}{0pt}{0pt}
    \multirow{2}{*}{Method} & \; & \multirow{2}{*}{Backbone} & \multirow{2}{*}{\makecell[c]{\#Trainable \\ Params}} & \multirow{2}{*}{Pre-Train Data} & & \multicolumn{5}{c}{COCO 2017val} \\
    & & & & & & AP$^{\mathrm{box}}$ & AP$^{\mathrm{box}}_{50}$ & \; & AP$^{\mathrm{mask}}$ & AP$^{\mathrm{mask}}_{50}$ \\
\specialrule{0.1em}{0pt}{0pt}
    {\color{gray} DyHead-T$^\dagger$~\cite{DyHead}} & & {\color{gray} Swin-T~\cite{Swin}} & {\color{gray} $\approx$100M} & {\color{gray} -} & & {\color{gray} 49.7} & {\color{gray} 68.0} & & {\color{gray} -} & {\color{gray} -} \\
    DyHead-T~\cite{DyHead} & & Swin-T~\cite{Swin} & $\approx$100M & O365 & & 43.6 & - & & - & - \\
    GLIP-T (B)~\cite{GLIP} & & Swin-T~\cite{Swin} & 232M & O365 & & 44.9 & 61.5 & & - & - \\
    GLIP-T (C)~\cite{GLIP} & & Swin-T~\cite{Swin} & 232M & O365,GoldG & & 46.7 & 63.4 & & - & - \\
    GLIP-T~\cite{GLIP} & & Swin-T~\cite{Swin} & 232M & O365,GoldG,CC3M,SBU & & 46.6 & 63.1 & & - & - \\
\specialrule{0.07em}{0pt}{0pt}
    {\color{gray} DINO$^\dagger$~\cite{DINO}} & & {\color{gray} Swin-T~\cite{Swin}} & {\color{gray} 49M} & {\color{gray} -} & & {\color{gray} 54.4} & {\color{gray} 72.9} & & {\color{gray} -} & {\color{gray} -} \\
    DINO~\cite{DINO} & & Swin-T~\cite{Swin} & 49M & O365 & & 46.2 & - & & - & - \\
    G-DINO-T~\cite{Grounding-DINO} & & Swin-T~\cite{Swin} & 172M & O365 & & 46.7 & - & & - & - \\
    G-DINO-T~\cite{Grounding-DINO} & & Swin-T~\cite{Swin} & 172M & O365,GoldG & & 48.1 & - & & - & - \\
    G-DINO-T~\cite{Grounding-DINO} & & Swin-T~\cite{Swin} & 172M & O365,GoldG,Cap4M & & 48.4 & 64.4 & & - & - \\
\specialrule{0.07em}{0pt}{0pt}
{\color{gray} ViTDet$^\dagger$~\cite{ViTDet}} & & {\color{gray} ViT-B~\cite{ViT}} & {\color{gray} 141M} & {\color{gray} -} & & {\color{gray} 54.0} & {\color{gray} 72.2} & & {\color{gray} 46.7} & {\color{gray} 69.8} \\
\rowcolor{orange!10}    Sambor (Ours) & & ViT-B~\cite{ViT} & 160M & O365 & & 47.3 & 64.7 & & 36.5 & 59.8 \\
\rowcolor{orange!10}    Sambor$^\bigstar$ (Ours) & & ViT-B~\cite{ViT} & 160M & O365 & & \textbf{48.6} & \textbf{66.1} & & \textbf{37.1} & \textbf{60.6} \\
\specialrule{0.1em}{0pt}{0pt}
\end{tabular}}
\end{center}
\caption{\textbf{Zero-shot transfer performance on COCO benchmark.} 
$^\bigstar$denotes the application of Open-set RPN. $^\dagger$denotes supervised approaches.}
\label{tab:coco}
\end{table*}

\setlength{\tabcolsep}{1.5pt}
\begin{table*}[tb]
\renewcommand\arraystretch{1.23}
\begin{center}
\resizebox{1\textwidth}{!}{
\begin{tabular}{lccccccccccccc}
\specialrule{0.1em}{0pt}{0pt}
    \multirow{2}{*}{Method} & \multirow{2}{*}{Backbone} & \multirow{2}{*}{\makecell[c]{\#Trainable \\ Params}} & \multirow{2}{*}{Pre-Train Data} & & \multicolumn{3}{c}{MiniVal} & & \multicolumn{3}{c}{Val v1.0} \\
    & & & & & AP & \: & AP$_{r}$\textbf{/}AP$_{c}$\textbf{/}AP$_{f}$ & \;\; & AP & \: & AP$_{r}$\textbf{/}AP$_{c}$\textbf{/}AP$_{f}$ \\
\specialrule{0.1em}{0pt}{0pt}
    {\color{gray} MDETR$^\dagger$~\cite{MDETR}} & {\color{gray} RN101} & {\color{gray} 186M} & {\color{gray} GoldG+,RefC} & & {\color{gray} 24.2} & & {\color{gray} 20.9}\textbf{/}{\color{gray} 24.9}\textbf{/}{\color{gray} 24.3} & & {\color{gray} -} & & {\color{gray} -} \textbf{/} {\color{gray} -} \textbf{/} {\color{gray} -} \\
    {\color{gray} Mask R-CNN$^\dagger$~\cite{MDETR}} & {\color{gray} RN101} & {\color{gray} 69M} & {\color{gray} -} & & {\color{gray} 33.3} & & {\color{gray} 26.3}\textbf{/}{\color{gray} 34.0}\textbf{/}{\color{gray} 33.9} & & {\color{gray} -} & & {\color{gray} -} \textbf{/} {\color{gray} -} \textbf{/} {\color{gray} -} \\
\specialrule{0.07em}{0pt}{0pt}
    GLIP-T (B)~\cite{GLIP} & Swin-T & 232M & O365 & & 17.8 & & 13.5\textbf{/}12.8\textbf{/}22.2 & & 11.3 & & 4.2\textbf{/}7.6\textbf{/}18.6 \\
    GLIP-T (C)~\cite{GLIP} & Swin-T & 232M & O365,GoldG & & 24.9 & & 17.7\textbf{/}19.5\textbf{/}31.0 & & 16.5 & & 7.5\textbf{/}11.6\textbf{/}26.1 \\
    GLIP-T~\cite{GLIP} & Swin-T & 232M & O365,GoldG,Cap4M & & 26.0 & & 20.8\textbf{/}21.4\textbf{/}31.0 & & 17.2 & & 10.1\textbf{/}12.5\textbf{/}25.5 \\
    GLIPv2-T~\cite{GLIPv2} & Swin-T & 232M & O365,GoldG,Cap4M & & 29.0 & & - \textbf{/} - \textbf{/} - & & - & & - \textbf{/} - \textbf{/} - \\
\specialrule{0.07em}{0pt}{0pt}
    DetCLIP-T (A)~\cite{DetCLIP} & Swin-T & - & O365 & & 28.8 & & 26.0\textbf{/}28.0\textbf{/}30.0 & & 22.1 & & 18.4\textbf{/}20.1\textbf{/}26.0 \\
    DetCLIP-T (B)~\cite{DetCLIP} & Swin-T & - & O365,GoldG & & 34.4 & & 26.9\textbf{/}33.9\textbf{/}36.3 & & 27.2 & & 21.9\textbf{/}25.5\textbf{/}31.5 \\
    DetCLIP-T~\cite{DetCLIP} & Swin-T & - & O365,GoldG,YFCC1M & & 35.9 & & 33.2\textbf{/}35.7\textbf{/}36.4 & & 28.4 & & 25.0\textbf{/}27.0\textbf{/}31.6 \\
    DetCLIPv2-T~\cite{DetCLIPv2} & Swin-T & - & O365 & & 28.6 & & 24.2\textbf{/}27.1\textbf{/}30.6 & & - & & - \textbf{/} - \textbf{/} - \\
    DetCLIPv2-T~\cite{DetCLIPv2} & Swin-T & - & O365,CC3M & & 31.3 & & 29.4\textbf{/}31.7\textbf{/}31.3 & & - & & - \textbf{/} - \textbf{/} - \\
    DetCLIPv2-T~\cite{DetCLIPv2} & Swin-T & - & O365,GoldG,CC3M & & 38.4 & & \textbf{36.7}\textbf{/}37.9\textbf{/}39.1 & & - & & - \textbf{/} - \textbf{/} - \\
\specialrule{0.07em}{0pt}{0pt}
    G-DINO-T~\cite{Grounding-DINO} & Swin-T & 172M & O365,GoldG & & 25.6 & & 14.4\textbf{/}19.6\textbf{/}32.2 & & - & & - \textbf{/} - \textbf{/} - \\
    G-DINO-T~\cite{Grounding-DINO} & Swin-T & 172M & O365,GoldG,Cap4M & & 27.4 & & 18.1\textbf{/}23.3\textbf{/}32.7 & & - & & - \textbf{/} - \textbf{/} - \\
\specialrule{0.07em}{0pt}{0pt}
\rowcolor{orange!10}   Sambor$^\bigstar$ (Ours) & ViT-B & 160M & O365 & & 33.1 & & 29.6\textbf{/}32.0\textbf{/}34.7 & & 26.3 & & 20.9\textbf{/}24.4\textbf{/}30.9  \\
\rowcolor{orange!10}   Sambor$^\bigstar$ (Ours) & ViT-B & 160M & O365,GoldG & & \textbf{39.6} & & 34.6\textbf{/}\textbf{39.3}\textbf{/}\textbf{40.7} & & \textbf{32.8} & & \textbf{30.9}\textbf{/}\textbf{31.0}\textbf{/}\textbf{35.7}  \\
\specialrule{0.1em}{0pt}{0pt}
\end{tabular}}
\end{center}
\caption{\textbf{Zero-shot object detection performance on LVIS benchmark.} AP$_r$, AP$_c$, and AP$_f$ indicate the AP values for rare, common, and frequent categories, respectively. $^\bigstar$denotes the application of Open-set RPN. $^\dagger$denotes supervised approaches.}
\label{tab:lvis}
\end{table*}

\setlength{\tabcolsep}{1.2pt}
\begin{table}[th!]
\renewcommand\arraystretch{1.23}
\begin{center}
\resizebox{\columnwidth}{!}{
\begin{tabular}{lccccccc}
\specialrule{0.1em}{0pt}{0pt}
    \multirow{2}{*}{Method} & & \multirow{2}{*}{Pre-Train Data} & & \multicolumn{2}{c}{MiniVal} & & \multicolumn{1}{c}{Val v1.0} \\
    & & & & AP & AP$_{r}$\textbf{/}AP$_{c}$\textbf{/}AP$_{f}$ & & AP \\
\specialrule{0.1em}{0pt}{0pt}
X-Decoder (T) & & COCO,C4M$^\ddagger$ & & - & - \textbf{/} - \textbf{/} - & & 9.6 \\
OpenSeeD (T) & & O365,COCO & & - & - \textbf{/} - \textbf{/} - & & 19.4 \\
OpenSeeD (L) & & O365,COCO & & - & - \textbf{/} - \textbf{/} - & & 21.0 \\
\specialrule{0.07em}{0pt}{0pt}
\rowcolor{orange!10}   Sambor$^\bigstar$ (Ours) & & O365 & & 27.6 & 27.3\textbf{/}27.5\textbf{/}27.7 & & 21.7 \\
\rowcolor{orange!10}   Sambor$^\bigstar$ (Ours) & & O365,GoldG & & \textbf{35.7} & \textbf{31.4}\textbf{/}\textbf{37.1}\textbf{/}\textbf{35.3} & & \textbf{29.4} \\
\specialrule{0.1em}{0pt}{0pt}
\end{tabular}}
\caption{\textbf{Zero-shot instance segmentation performance on LVIS benchmark.} AP$_r$, AP$_c$, and AP$_f$ indicate the AP values for rare, common, and frequent categories, respectively. $^\ddagger$\,includes Conceptual Captions~\cite{CC3M}, SBU Captions~\cite{SBU}, 
Visual Genome~\cite{VG}, and COCO Captions~\cite{COCO-Captions}. $^\bigstar$denotes the application of Open-set RPN.}
\label{tab:lvis-segm}
\end{center}
\end{table}

\subsection{Open-Vocabulary Classification}
\label{sec:classification}
The second stage of the detector focuses on transforming the proposals into a set of predicted bounding boxes $B=\{\mathbf{b}_{k}\}_{k=1}^{K}$ ($K$ is the number of predictions) along with the classification features $\mathcal{F}^{B}\in\mathbb{R}^{K\times D}$.
To expand the classification into open vocabulary, we introduce a language branch, \ie, the CLIP text encoder.
We insert each concept name into the prompt template to form a complete sentence. These sentences are separately forwarded to the text encoder for obtaining sentence embeddings $\mathcal{F}^{T}\in\mathbb{R}^{M\times D}$, where $M$ is the number of concepts sampled in each batch.
We calculate the similarity matrix $S=\mathcal{F}^{B}\cdot(\mathcal{F}^{T})^\top\in\mathbb{R}^{K\times M}$ to construct the word-region alignment loss~\cite{GLIP}.
In our design, text embeddings are solely utilized to predict word-region similarity scores without the additional cross-modal fusion adopted in~\cite{GLIP,Grounding-DINO}.

\noindent
\textbf{Unified Data Formulation.}
Following~\cite{DetCLIP,DetCLIPv2}, we employ a paralleled formulation to unify the data formats from object detection and phrase grounding.

\begin{itemize}[left=0pt,labelsep=3pt,labelwidth=1em,align=left]
\item \textbf{Object Detection.} 
A concept set, based on dataset category names, designates present categories as positives and absent ones as negatives.
Given that the category names in the dataset remain unchanged, extracting and storing category features before training can avoid redundant feature extraction and improve efficiency.
Additionally, each category feature is the average of all its prompt templates, a policy applied during both training and inference.
\item \textbf{Phrase Grounding.}
We extract phrases corresponding to labeled objects from the caption to form a positive concept set.
We utilize the large-scale and information-dense Bamboo~\cite{bamboo} to diversify negative concepts.
Given the variability of concept names in images, we dynamically select a subset of concepts as negatives for each batch, setting the total number of positive and negative concepts to 150. For efficiency, we randomly choose a prompt template to extract text features.
\end{itemize}

\section{Experiments}
\subsection{Implementation Details}

\textbf{Training Datasets.}
For object detection, we use the Objects365~\cite{Objects365} dataset (referred to as O365), comprising 365 categories. For phrase grounding, we use the GoldG~\cite{MDETR} dataset, which contains well-annotated images from sources including Flickr30K~\cite{Flickr30k}, VG Caption~\cite{VG}, and GQA~\cite{GQA}. We deliberately exclude COCO~\cite{COCO} images to ensure a more equitable evaluation of zero-shot transfer performance.

\noindent
\textbf{Training Details.}
We pre-train our models using SAM with ViT-B~\cite{ViT} as the backbone and CLIP with RN50$\times$64~\cite{resnet}, using a batch size of 64. We select AdamW~\cite{adamw} optimizer with a 0.05 weight decay, an initial learning rate 4$\times$10$^{-4}$, and a cosine annealing learning rate decay. The default training schedule is 12 epochs. 
The input image size is 1,024$\times$1,024 with standard scale jittering~\cite{SSJ}. This size is uniformly employed as the input for SAM, CLIP, and SideFormer. The max token length for each input sentence follows the CLIP default setting of 77. MMDetection~\cite{mmdetection} code-base is used.

\subsection{Zero-Shot Transfer Performance}
\noindent
\textbf{COCO Benchmark}~\cite{COCO}, comprising 80 common object categories, stands as the most widely utilized dataset for object detection.
Considering that O365 covers all 80 categories and is frequently employed as pre-training data for COCO, we focus on evaluating the zero-shot transfer performance for models pre-trained with O365.

We provide a comparison between GLIP~\cite{GLIP} and Grounding DINO (G-DINO)~\cite{Grounding-DINO}, along with their underlying detectors 
in Table~\ref{tab:coco}. Our approach outperforms previous methods on the zero-shot transfer settings. When pre-trained on the same O365 dataset, Sambor shows \textbf{$+$2.4}~AP and \textbf{$+$0.6}~AP compared to GLIP and G-DINO, respectively.
When employing the Open-set RPN (see more details in Sec.~\ref{sec:ablation-rpn}), Sambor demonstrates the best performance, surpassing even models that utilize larger datasets.
We additionally report the zero-shot instance segmentation performance, which is effortlessly achievable by feeding the detection outputs into the SAM head.
Notably, in comparison to other models, Sambor has a lower count of trainable parameters. This is attributed to the fact that, aside from SideFormer and the detection head, the parameters of the remaining components are frozen.

\noindent
\textbf{LVIS Benchmark}~\cite{LVIS} contains 1,203 categories, including numerous rare categories that are seldom encountered in pre-training datasets. We report the \emph{Fixed} AP~\cite{fixed-ap} on both the MiniVal~\cite{MDETR} subset, comprising 5,000 images, and the complete validation set v1.0.

The zero-shot transfer performance on LVIS is presented in Table~\ref{tab:lvis}. 
Here, we also report the performance with and without the use of Open-set RPN, revealing an improvement of 0.4 AP when employed. Under comparable volumes of training data, Sambor outperforms competitors by a large margin. Specifically, in the scenario of training solely on O365 and evaluating on LVIS MiniVal, our model outperforms GLIP by \textbf{15.3} AP, and surpasses DetCLIP\textbf{/}DetCLIPv2 by \textbf{4.3/4.5} AP, respectively. The advantage is similarly pronounced when compared to G-DINO.
Further, we incorporate the phrase grounding dataset GoldG to facilitate the generalization of Sambor. Instead of training from scratch, we fine-tune the O365 pre-trained model for 3 epochs while keeping other settings consistent.
Evidently, Sambor exhibits enhanced zero-shot transfer performance across all categories. It surpasses even the results achieved with larger datasets compared to previous methods.
Moreover, we present the zero-shot instance segmentation performance in Table~\ref{tab:lvis-segm}, where the mask results are generated by prompting the detection boxes to the SAM head.
Compared to X-Decoder~\cite{X-Decoder} and OpenSeeD~\cite{OpenSeeD}, our Sambor exhibits superior performance, indicating its robust zero-shot generalization.

\subsection{Ablation Studies}
\label{sec:ablation}
We conduct a series of ablation studies on Sambor, training with default settings on O365 unless specified otherwise.

\noindent
\textbf{Effectiveness of SideFormer.}
Table~\ref{tab:ablation-sideformer} validates the effectiveness of our designed SideFormer on COCO and LVIS. The Sambor baseline represents directly connecting the ViTDet to the SAM image encoder.
We first incorporate the SAM Extractor to obtain multi-level features and perform fine-tuning. 
To further enhance the model's recognition capability, we introduce the CLIP Injector to improve semantic representation, achieving optimal performance. Furthermore, we utilize region-word alignment for open-vocabulary classification, but there is a noticeable domain gap between CLIP text features and region features. Consequently, incorporating CLIP visual features effectively bridges this gap and provides benefits for Sambor.

\setlength{\tabcolsep}{2.5pt}
\begin{table}[t]
\renewcommand\arraystretch{1.2}
\begin{center}
\resizebox{\columnwidth}{!}{
\begin{tabular}{lccccccccc}
\specialrule{0.1em}{0pt}{0pt}
    \multirow{2}{*}{Strategy} & & \multicolumn{3}{c}{COCO val} & & \multicolumn{4}{c}{LVIS MiniVal} \\
    & & AP & AP$_{50}$ & AP$_{75}$ & & AP & AP$_{r}$ & AP$_{c}$ & AP$_{f}$ \\
\specialrule{0.1em}{0pt}{0pt}
    Sambor baseline & & 39.0 & 54.7 & 42.5 & & 27.7 & 21.5 & 26.5 & 29.9  \\
    $+$ SAM Extractor & & 42.4 & 58.7 & 46.2 & & 29.0 & 25.2 & 27.5 & 31.0 \\
    $+$ CLIP Injector & & \textbf{47.3} & \textbf{64.7} & \textbf{51.3} & & \textbf{32.7} & \textbf{29.5} & \textbf{32.1} & \textbf{33.9} \\
\specialrule{0.1em}{0pt}{0pt}
\end{tabular}}
\end{center}
\caption{\textbf{Ablation studies on the components in SideFormer.} The combination of SAM Extractor and CLIP Injector shows the best performance.}
\label{tab:ablation-sideformer}
\end{table}

\setlength{\tabcolsep}{1.3pt}
\begin{table}[t]
\renewcommand\arraystretch{1.2}
\begin{center}
\resizebox{\columnwidth}{!}{
\begin{tabular}{lcccccc}
\specialrule{0.1em}{0pt}{0pt}
    \multirow{2}{*}{Strategy} & & \multicolumn{2}{c}{COCO val} & & \multicolumn{2}{c}{LVIS MiniVal} \\
    & & \multicolumn{1}{c}{AR${@1000}$} & \multicolumn{1}{c}{AP} & & \multicolumn{1}{c}{AR${@1000}$} & \multicolumn{1}{c}{AP} \\
\specialrule{0.1em}{0pt}{0pt}
    Vanilla RPN 
    & & 65.4 & 47.3 & & 49.1 & 32.7 \\
    Open-set RPN\,$^\sharp$ & & \textbf{67.5 \footnotesize (+2.1)} & 46.7 \footnotesize (-0.6) & & \textbf{54.8 \footnotesize (+5.7)} & 29.3 \footnotesize (-3.4) \\
    Open-set RPN & & \textbf{67.5 \footnotesize (+2.1)} & \textbf{48.6 \footnotesize (+1.3)} & & \textbf{54.8 \footnotesize (+5.7)} & \textbf{33.1 \footnotesize (+0.4)} \\
\specialrule{0.1em}{0pt}{0pt}
\end{tabular}}
\end{center}
\caption{\textbf{Ablation studies on Open-set RPN.} Without further fine-tuning using additional region proposals from Open-set RPN (denoted as $^\sharp$), the improvement in proposal quality (AR${@1000}$) brought by Open-set RPN cannot be directly translated to an increase in detection performance (AP). Fine-tuning serves as a remedy for this discrepancy, yielding superior results.}
\label{tab:proposal}
\end{table}

\setlength{\tabcolsep}{2.pt}
\begin{table}[t]
\renewcommand\arraystretch{1.2}
\begin{center}
\resizebox{\columnwidth}{!}{
\begin{tabular}{lccccccccc}
\specialrule{0.1em}{0pt}{0pt}
    Proposal & AR${@1000}$ & & AP & AP$_{50}$ & AP$_{75}$ & & AP$_s$ & AP$_m$ & AP$_l$ \\
\specialrule{0.1em}{0pt}{0pt}
    \emph{only} RPN & 65.4 & & 47.3 & 64.7 & 51.3 & & \textbf{33.5} & \textbf{53.2} & 61.4 \\
    \emph{only} SAM & 59.4 & & 46.8 & 63.3 & 51.1 & & 29.8 & 52.2 & \textbf{65.0} \\
    Open-set RPN & \textbf{67.5} & & \textbf{48.6} & \textbf{66.1} & \textbf{52.7} & & \textbf{33.5} & \textbf{53.2} & 64.2 \\
\specialrule{0.1em}{0pt}{0pt}
\end{tabular}}
\end{center}
\caption{\textbf{Ablation studies on region proposal sources} for COCO 2017val. Using region proposals only from SAM exhibits a noticeable performance gap, particularly for small objects. Open-set RPN can effectively combine two sets of proposals, thereby achieving optimal performance.}
\label{tab:rpn}
\end{table}

\noindent
\textbf{Effectiveness of Open-set RPN.}
\label{sec:ablation-rpn}
As elaborated in Sec.~\ref{sec:proposal}, the automatic mask generation pipeline of SAM allows for producing a number of high-quality open-set proposals, serving as a valuable complement to the vanilla RPN.
We use a 32$\times$32 grid of points to generate open-set proposals, with a post-processing NMS threshold set to 0.7.
Table~\ref{tab:proposal} illustrates the average recall (AR${@1000}$) for proposals, and it is evident that incorporating these open-set proposals significantly improves AR.
However, the improved quality of region proposals does not manifest as superior detection performance; instead, there has been a decline. We posit that this is attributed to the detection head in the second stage not being exposed to the additional region proposals during training, hindering the ability to process them effectively.

To address this issue, we conduct a minor-scale fine-tuning adopting the Open-set RPN, \ie, incorporating these open-set proposals during training. Specifically, with considerations for training efficiency, we use a 32$\times$32 grid of points to fine-tune for 1 epoch on approximately one-fifth of the O365 dataset.
Maintaining all other hyper-parameters constant, employing a reduced learning rate of 4$\times$10$^{-5}$ contributes to the efficacy of fine-tuning.
It effectively eradicates inconsistencies in results, leading to superior performance for the Open-set RPN. Compared to the vanilla RPN, there is an improvement of 1.3 AP on COCO and 0.4 AP on LVIS. Unless specified otherwise, the Open-set RPN described elsewhere in this paper is fine-tuned.

\noindent
\textbf{Region Proposal Sources.}
After establishing the effectiveness of the Open-set RPN, a relevant question arises: \emph{What would be the impact if we solely rely on object proposals from SAM?}
We conduct ablation studies on the model fine-tuned with Open-set RPN. The evaluations include performance using proposals only from RPN, proposals only from the SAM head, and a combination of both sets, as shown in Table~\ref{tab:rpn}. 
Only RPN performance remains consistent compared to before fine-tuning (first row in Table~\ref{tab:proposal}). This confirms that the performance improvement in Open-set RPN is attributed to the supplementary proposals rather than the fine-tuning impact on RPN.
To ensure an adequate quantity when relying solely on proposals from the SAM head, we increase the density of grid points to 64$\times$64 and set the NMS threshold to 0.95. There is a noticeable decrease in detection performance, especially for small objects. The difficulty of precisely targeting small objects with sampling points contributes to the inability to recall them, aligning with the results shown in~\citet{SAM}. Moreover, continuously increasing the density is impractical as it introduces unbearable computational and time consumption. Hence, the integration of the trainable RPN is crucial. The Open-set RPN effectively combines the two sets of region proposals in a complementary fashion, achieving optimal performance.

\begin{figure*}[t]
  \centering
  \includegraphics[width=1\linewidth]{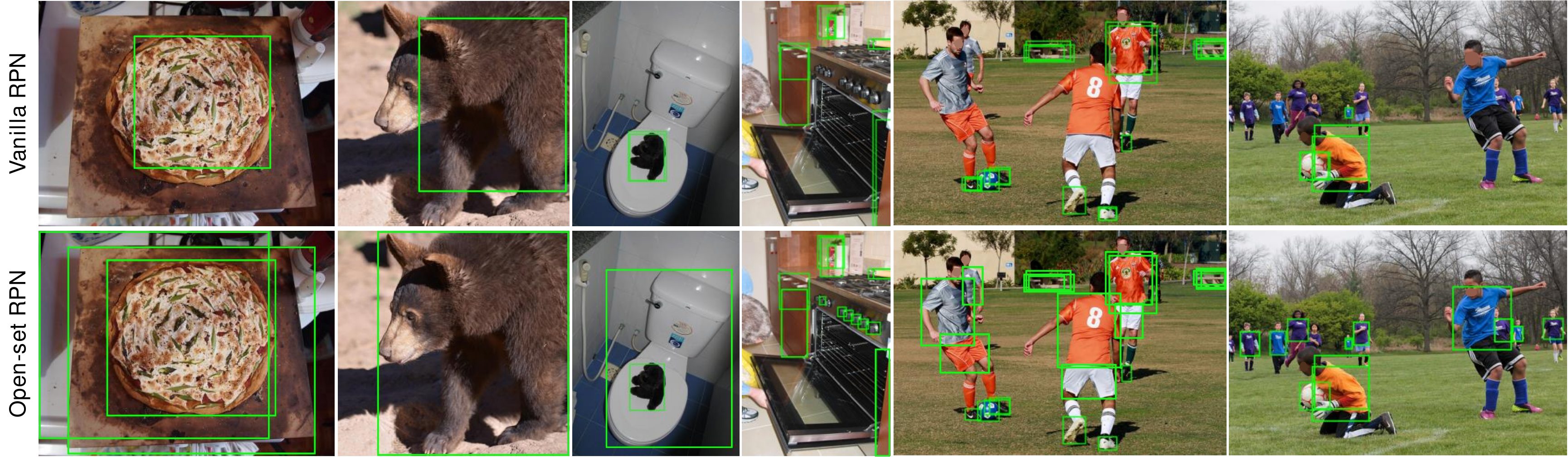}
  \caption{\textbf{Visualization comparison between Open-set RPN and the vanilla RPN.} For clarity, we only display high-quality proposals with an IoU greater than 0.7 with the ground truth boxes. In the first two examples, the vanilla RPN fails to generate  proposals meeting this criterion; thus, we show the one with the highest IoU.}
  \label{fig:vis-rpn}
\end{figure*}

\begin{figure*}[t]
  \centering
  \includegraphics[width=1\linewidth]{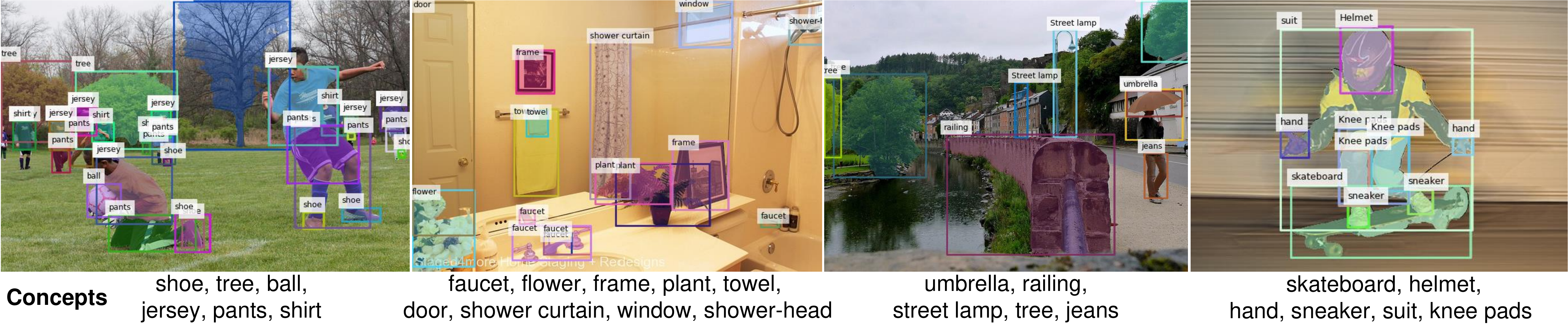}
  \caption{\textbf{Visualization of Sambor for open-vocabulary object detection and instance segmentation.} For better mask visual effects, we adopt HQ-SAM~\cite{HQ-SAM} as the mask decoder.}
  \label{fig:vis-result}
\end{figure*}

\subsection{Analyses}
\label{sec:analyses}
\noindent
\textbf{Open-Vocabulary Object Detector Built upon SAM.}
The release of SAM has generated significant interest within the community, leading to the development of numerous derivative models. 
Instead of merely combining with SAM in a cascade manner, our Sambor seamlessly integrates SAM with an open-vocabulary detector into a unified end-to-end framework. This not only enables feature sharing but also facilitates the efficiency of interactive operations within the system.
The advantages of Sambor are evident in the mutually beneficial relationship between SAM and the open-vocabulary object detector. (1) SAM provides the detector with powerful generalization features, yielding competitive zero-shot performance after semantic information supplementation. Moreover, the Open-set RPN leveraging proposals generated by SAM further enhances the object recall in open-world scenarios. (2) The open-vocabulary object detector endows SAM to recognize arbitrary objects. Consequently, when utilizing functionalities from SAM, \eg, interactive prompts, the detector can predict the categories while outputting segmentation results. This facilitates a more effortless and accurate acquisition of the desired targets.

\noindent
\textbf{Visualization of Open-set RPN.}
The comparison of object proposals between the Open-set RPN and the vanilla RPN is illustrated in Fig.~\ref{fig:vis-rpn}. Thanks to incorporating more generalized proposals, the Open-set RPN effectively compensates for the recall of some objects, thereby addressing more challenging scenarios. For example, it can accurately capture the overall edges of an object and effectively leverage color information to delineate a person's clothing, areas where the vanilla RPN performs inadequately.

\noindent
\textbf{Visualization of Sambor.}
\label{sec:analyses-rpn}
We present the open-vocabulary object detection results of Sambor in Fig.~\ref{fig:vis-result}, where the desired objects are detected by inputting category concepts. Additionally, we feed the detection boxes into the SAM head to obtain instance segmentation masks.

\section{Conclusion}
Sambor is an end-to-end open-vocabulary object detector that integrates the vision foundation model SAM. It effectively leverages the SAM features along with class-agnostic proposals to boost object detection. On the other hand, the open-vocabulary object detector supplements SAM with the lacking classification capability, thus empowering it to extend beyond segmenting anything to recognizing arbitrary categories.
Experiments demonstrate the superior open-vocabulary performance of Sambor and the contributions of the proposed modules.
We aim for this effort to be an effective attempt, equipping SAM with recognition capabilities to address a wider array of application needs and offering a potential direction for open-vocabulary learning.

\section*{Acknowledgements}
This work was supported in part by the Key Deployment Program of the Chinese Academy of Sciences under Grant KGFZD145-23-18 and the Strategic Priority Research Program of the Chinese Academy of Sciences under Grant E1XA310103.

\setcounter{section}{0}
\renewcommand\thesection{\Alph {section}}

\section{More Implementation Details}
\textbf{Pre-training details.} 
Table~\ref{tab:train-detail} summarizes the detailed settings we use for pre-training and fine-tuning. In addition, for the Cascade R-CNN~\cite{Cascade} detector head, we employ a modified Focal Loss~\cite{focalloss} (adapted in the form of word-region alignment) for classification, and the SmoothL1 Loss~\cite{fast-rcnn} is implemented for regression.

\noindent
\textbf{Text prompts.}
Since the vision-language foundation model is trained with complete sentences, we input the concepts into a prompt template and use an ensemble of diverse prompts.
We utilize a list of 80 prompt templates following~\cite{CLIP} without bells and whistles.

\setlength{\tabcolsep}{5pt}
\begin{table}[t]
\renewcommand\arraystretch{1.1}
\begin{center}
\resizebox{\columnwidth}{!}{
\begin{tabular}{llll}
\specialrule{0.1em}{0pt}{0pt}
Configuration & & & Value \\
\specialrule{0.1em}{0pt}{0pt}
\multicolumn{4}{l}{\textbf{\emph{Pre-training:}}}  \\
Dataset & & & O365 \\
SAM Backbone & & & ViT-B \\
CLIP Backbone & & & RN50$\times$64 \\
Training Epochs & & & 12 \\
Batch Size & & & 64 \\
Optimizer & & & AdamW \\
LR & & & 4$\times$10$^{-4}$ \\
LR Schedule & & & CosineAnnealing \\
Weight Decay & & & 0.05 \\
Warmup Iters & & & 1,000 \\
Image Resolution & & & 1,024$\times$1,024 \\
Augmentation & & & Random Flip, SSJ~\shortcite{SSJ} \\
Max Text Token Length & & & 77 \\
SideFormer Attention & & & DeformableAttention~\shortcite{deformable} \\
\# of Attention Heads & & & 8 \\
Feedforward Channels & & & 3072 \\
\specialrule{0.1em}{0pt}{0pt}
\multicolumn{4}{l}{\textbf{\emph{Open-set RPN Fine-tuning:}}}  \\
Dataset & & & $\approx \nicefrac {1}{5} \;$O365 \\
LR & & & 4$\times$10$^{-5}$ \\
Training Epochs & & & 1 \\
\# of Grid Points & & & 32$\times$32 \\
RPN NMS Threshold & & & 0.7 \\
\specialrule{0.1em}{0pt}{0pt}
\multicolumn{4}{l}{\textbf{\emph{Unified Data Fine-tuning:}}}  \\
Dataset & & & O365 $+$ GoldG \\
LR & & & 4$\times$10$^{-4}$ \\
Training Epochs & & & 3 \\
\# of Concepts$^\dagger$ & & & 150 \\
\specialrule{0.1em}{0pt}{0pt}
\end{tabular}}
\caption{Detailed training settings of Sambor. $^\dagger$only for phrase grounding data.}
\label{tab:train-detail}
\end{center}
\end{table}





\bibliography{aaai25}

\end{document}